\documentclass[letterpaper]{article} 
\usepackage{aaai2027}  
\usepackage[hyphens]{url}  
\usepackage{graphicx} 
\usepackage{amsmath}
\usepackage{natbib}  
\usepackage{caption} 
\nocopyright
\usepackage{algorithm}
\usepackage{algorithmic}

\usepackage{newfloat}
\usepackage{listings}
\DeclareCaptionStyle{ruled}{labelfont=normalfont,labelsep=colon,strut=off} 
\floatstyle{ruled}
\newfloat{listing}{tb}{lst}{}
\floatname{listing}{Listing}

\usepackage{booktabs}

\title{HexEval: An Evidence-Driven Hexagonal Framework for Multidimensional Scholar Assessment}

\author{
    Xiaokang Qu\textsuperscript{\rm 1},
    Yiting Lin\textsuperscript{\rm 1}
}

\affiliations{
    \textsuperscript{\rm 1}School of Cyber Science and Technology, 
    University of Science and Technology of China\\
    Hefei, Anhui, China\\
    xkqu@mail.ustc.edu.cn, linyiting@mail.ustc.edu.cn
}

\begin{document}

\maketitle

\begin{abstract}

Scholar assessment plays a fundamental role in faculty recruitment, funding allocation, academic promotion, and talent discovery. Existing scholar assessment methods predominantly rely on bibliometric indicators and reputation proxies, while recent large language model (LLM)-based approaches mainly focus on evaluating individual research papers rather than comprehensively assessing scholars. We argue that scholar assessment should be formulated as an \emph{evidence-driven reasoning} problem that jointly considers intrinsic research quality and externally verifiable scholarly behavior. To this end, we propose \textbf{HexEval}, an evidence-driven hexagonal framework for multidimensional scholar assessment. HexEval explicitly organizes scholar assessment into two complementary evidence layers. The intrinsic layer evaluates anonymized representative works along three dimensions, namely research rigor, methodological innovation, and scientific contribution, whereas the external layer characterizes scholars through knowledge translation, research coherence, and academic impact using heterogeneous evidence collected from GitHub, Lens, OpenAlex, and other publicly verifiable sources. Instead of producing opaque aggregate scores, HexEval preserves intermediate evidence, dimension-specific rationales, and verification signals throughout the evaluation process, enabling interpretable and auditable scholar profiles. Experiments across all six dimensions show dimension-dependent agreement with human or external reference criteria: structured calibration improves absolute agreement for intrinsic quality, while the external modules recover broad trajectory and ordinal impact signals. These results support evidence-driven reasoning over heterogeneous scholarly evidence as a promising paradigm for auditable AI-assisted scholar assessment, while exposing the coverage and attribution limitations of public scholarly data.

\end{abstract}


\section{Introduction}

Scholar assessment supports faculty recruitment, funding allocation, academic promotion, award nomination, and talent discovery. Existing methods predominantly rely on bibliometric indicators such as citation counts, publication numbers, h-index, field-normalized metrics, and publication venues \cite{hirsch2005index,hicks2015leiden}. Although useful for measuring scholarly visibility and influence, these indicators mainly capture research outcomes rather than intrinsic research quality. They are also affected by field-specific citation practices, academic age, venue prestige, cumulative advantage, and reputation effects \cite{merton1968matthew,hicks2015leiden}.

\begin{figure}[t]
    \centering
    \includegraphics[width=\columnwidth]{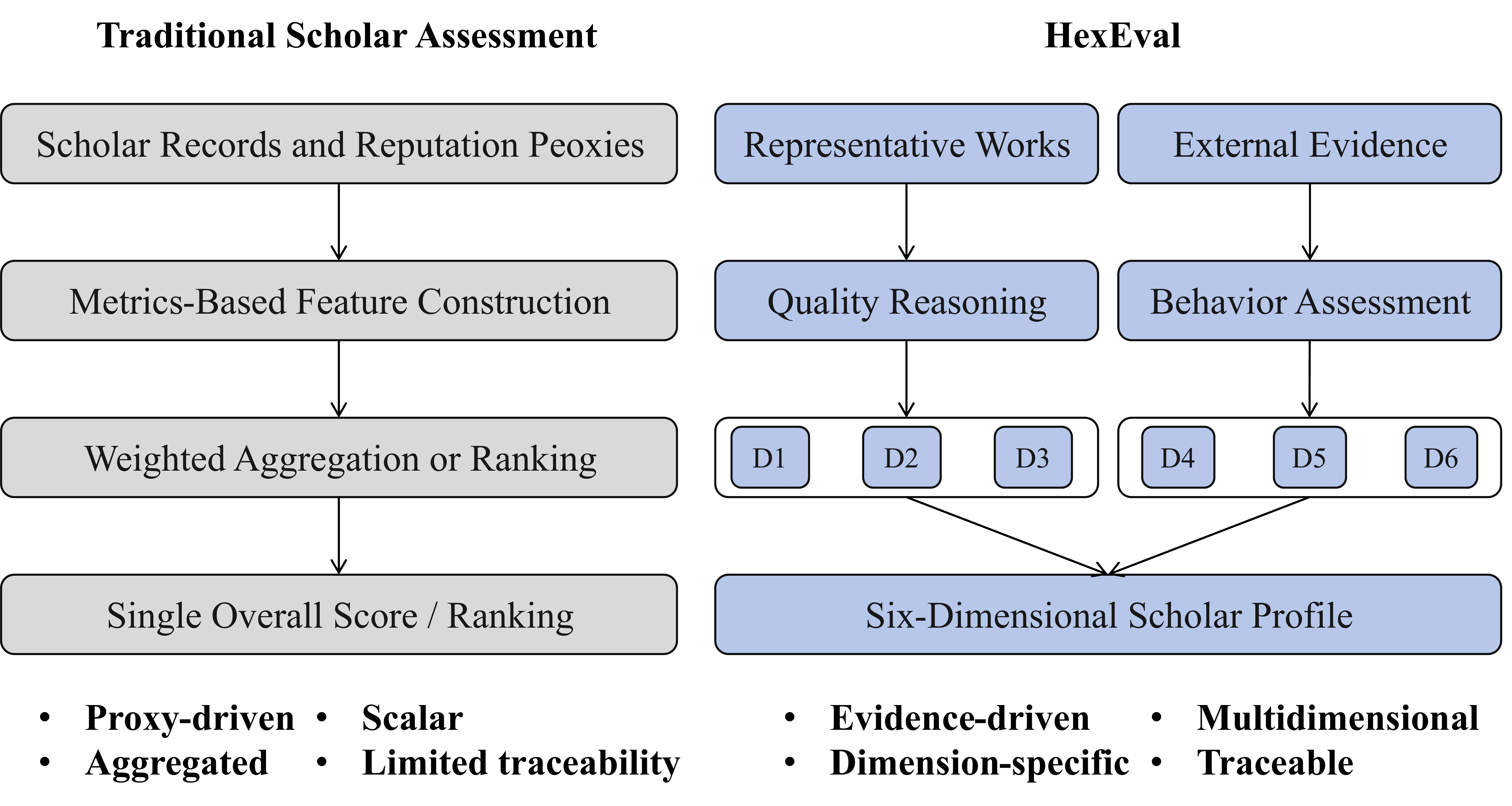}
    \caption{Paradigm comparison between conventional scholar assessment and HexEval. Conventional approaches aggregate metric-based features into a single score or ranking, whereas HexEval evaluates intrinsic research quality and external scholarly evidence separately to produce a traceable six-dimensional profile.}
    \label{fig:paradigm_comparison}
\end{figure}

Recent large language models (LLMs) have enabled structured scientific-document understanding and paper-level quality assessment with encouraging agreement with human judgments \cite{thelwall2024chatgpt,thelwall2025evaluating}. However, scholar assessment requires reasoning over broader evidence, including representative works, long-term research trajectories, knowledge translation, and scholarly impact. Paper-level content reasoning and scholar-level metric aggregation therefore remain largely disconnected.

We reformulate scholar assessment as a dual-layer evidence reasoning problem. As illustrated in Figure~\ref{fig:paradigm_comparison}, the intrinsic layer evaluates whether representative research is rigorous, innovative, and scientifically valuable, while the external layer characterizes how research is translated, sustained, and recognized in the scholarly ecosystem. This separation distinguishes scientific merit from downstream influence and produces more interpretable evaluation results.

Based on this formulation, we propose \textbf{HexEval}, an evidence-driven hexagonal framework that represents each scholar through six dimensions. The intrinsic layer evaluates anonymized representative works in terms of research rigor, methodological innovation, and scientific contribution. The external layer evaluates knowledge translation, research coherence, and academic impact using publicly verifiable evidence from GitHub, Lens, and OpenAlex \cite{priem2022openalex}. Each dimension uses its own evidence source, scoring procedure, and evaluation protocol, while preserving intermediate evidence, rationales, and verification signals for auditing.

The main contributions are:

\begin{itemize}
\item We formulate automated scholar assessment as a dual-layer evidence reasoning problem that separates intrinsic research quality from externally verifiable scholarly behavior.

\item We propose \textbf{HexEval}, a six-dimensional framework that independently evaluates research rigor, methodological innovation, scientific contribution, knowledge translation, research coherence, and academic impact while preserving auditable evidence.

\item We evaluate D1--D5 on public or curated reference data and operationalize D6 using the reproducible OpenAlex h-index, with explicit reporting of evidence coverage, attribution, sampling, and source limitations.
\end{itemize}

\section{Related Work}

\subsection{Scholar-Level Research Assessment}

Scholar-level assessment traditionally relies on publication counts, citations, the $h$-index, and field-normalized measures \cite{hirsch2005index,radicchi2008universality}; later work adds topics, authorship, time, collaboration, venue, and expert interpretation \cite{xie2021network,gorraiz2016individual}.

These indicators capture productivity and accumulated visibility more directly than intrinsic quality, and are affected by field practices, career length, coverage, authorship, and cumulative advantage \cite{merton1968matthew,hicks2015leiden}. Their association with peer judgment is field-dependent \cite{wainer2013correlations,thelwall2023fields}; responsible-assessment guidelines therefore treat them as context rather than substitutes for qualitative evidence \cite{hicks2015leiden,pontika2022indicators}.

\subsection{LLM-Based Research Evaluation}

LLMs now assist peer review through critique generation, methodological diagnosis, score prediction, and review improvement. Their plausible outputs remain limited by long-document understanding, paper-specific criticism, technical error detection, and score reliability \cite{zhou2024reliable,du2024critique}. Structured rubrics, retrieval, multi-stage reasoning, and agents improve consistency \cite{jin2024agentreview,zhu2025deepreview}, but deployment evidence favors reviewer assistance over autonomous decisions \cite{thakkar2025canlf}.

Direct quality estimation shows weak-to-moderate and field-dependent agreement with human judgments; repeated sampling, prompt design, and input selection materially affect results \cite{thelwall2024chatgpt,thelwall2025evaluating,thelwall2024inwhich}. Related work also considers research environments and societal value \cite{kousha2025chatgpt,nunkoo2026globalsouth}.

Existing methods remain output-centered and do not jointly represent representative-work quality, research trajectories, knowledge translation, and impact. HexEval instead treats the scholar as the target and preserves dimension-specific evidence traces.

\subsection{Evidence-Grounded Assessment}

Evidence-grounded methods support judgments with inspectable information and sources. Retrieval-augmented generation connects LLMs to external knowledge \cite{lewis2020retrieval}, while evidentiality-guided generation models evidence relevance and support \cite{asai2022evidentiality}; retrieval alone, however, does not guarantee valid support.

Scientific claim verification combines retrieval, support/refutation classification, and rationale extraction, as in SciFact and its extensions \cite{wadden2020fact,wadden-etal-2022-scifact}. Attribution-oriented methods such as RARR retain source links while revising unsupported claims \cite{gao2023rarr}. These works establish evidence--claim alignment and attribution as requirements for auditability \cite{jacovi2020faithfully}.

These methods mainly address QA, factual generation, or claim/document verification. Scholar assessment requires reasoning over works, career stages, artifacts, and impact channels; HexEval extends evidence-grounded reasoning to this setting with inspectable papers, software/patent records, trajectories, and citation traces.

\section{HexEval Framework}
\label{sec:framework}

\subsection{Framework Overview and Data Flow}

HexEval represents a scholar by scientific outputs and externally observable scholarly traces. Let $s$ denote a scholar, $\mathcal{P}_s$ the supplied representative works, and $\mathcal{I}_s$ the identity metadata required by attribution-dependent dimensions. The framework produces

\begin{equation}
\mathbf{H}(s)=
\left[D_1(s),D_2(s),D_3(s),D_4(s),D_5(s),D_6(s)\right],
\end{equation}

The first three dimensions measure intrinsic research quality and the last three measure externally observable scholarly behavior. Intrinsic scores use anonymized works; external scores use identity-linked evidence. The two paths are therefore complementary but not substitutable.

\begin{figure*}[t]
\centering
\includegraphics[width=0.8\textwidth]{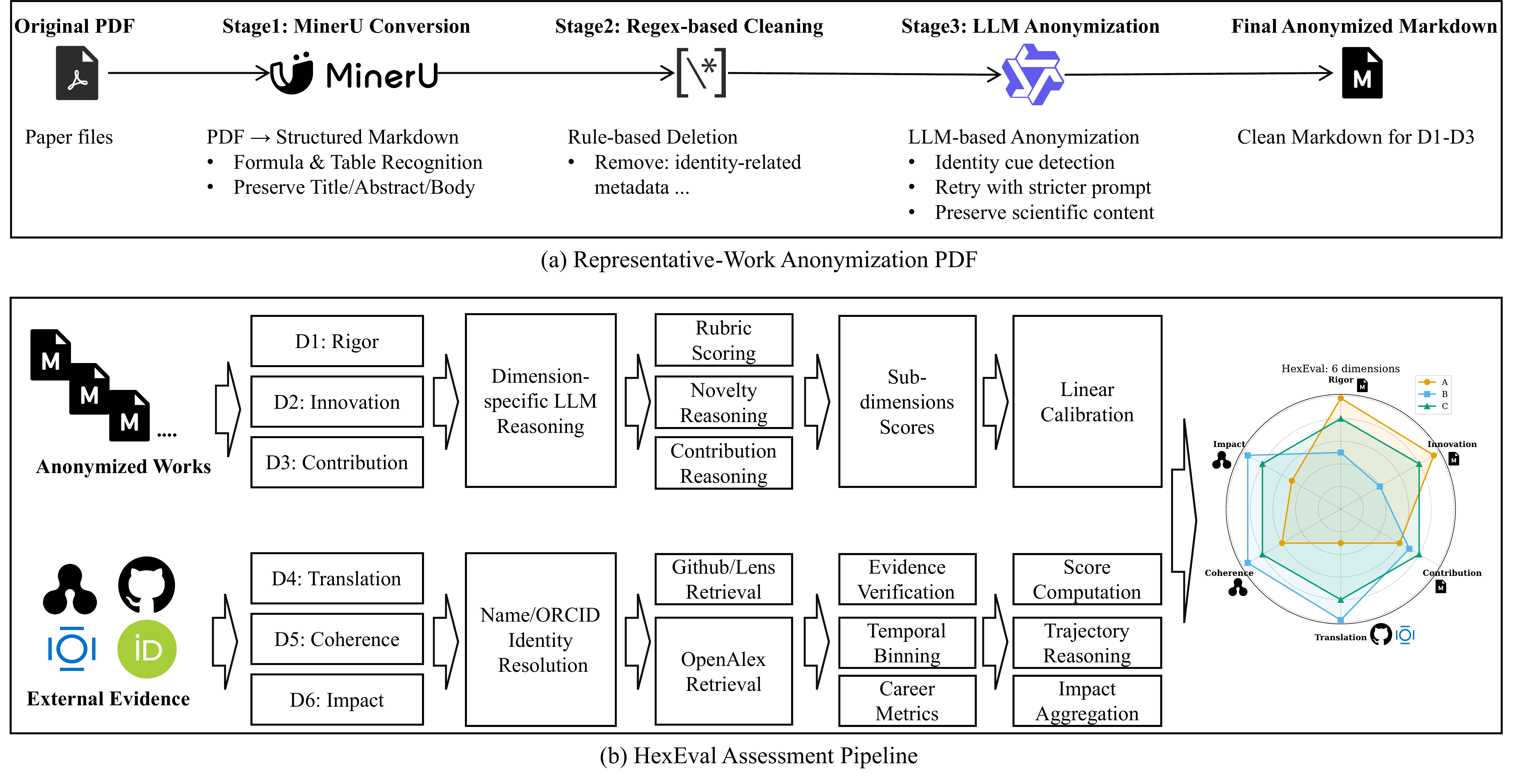}
\caption{Overview of the HexEval framework. (a) Representative-work anonymization pipeline, where original papers are converted into structured Markdown and processed through rule-based filtering and LLM-based identity cue removal to produce anonymized representative works. (b) Dual-path scholar assessment pipeline, where anonymized works are evaluated through three intrinsic dimensions (D1--D3), while external scholarly evidence is analyzed through three external dimensions (D4--D6). Each dimension independently produces evidence-grounded scores and rationales, which are integrated into an interpretable six-dimensional scholar profile. The radar chart provides illustrative scholar profiles rather than experimental results.}
\label{fig:hexeval_framework}
\end{figure*}

As summarized in Figure~\ref{fig:hexeval_framework}, each dimension has its own input schema, prompt or deterministic computation, score scale, and evidence record. HexEval does not impose a universal weighted sum; it returns the profile and evidence package

\begin{equation}
\begin{aligned}
\mathcal{O}(s)=\{&\mathbf{H}(s),\mathcal{E}_1(s),\ldots,\mathcal{E}_5(s),\\
&\mathcal{E}_6(s),\mathcal{V}(s)\},
\end{aligned}
\end{equation}

where $\mathcal{E}_i(s)$ contains inputs, structured outputs, rationales, and source metadata, and $\mathcal{V}(s)$ contains validation, coverage, and uncertainty information. Any downstream aggregation is application-specific and is not part of the default output.

For visualization only, a dimension score can be mapped from its native scale $[l_i,u_i]$ to a percentage:

\begin{equation}
\bar{D}_i(s)=100\,\frac{D_i(s)-l_i}{u_i-l_i},
\qquad l_i\leq D_i(s)\leq u_i.
\end{equation}

This affine mapping does not make the dimensions commensurate in a substantive sense and does not define a global scholar ranking. We use the notation

\begin{equation}
\begin{aligned}
L(x;c)&=\min\!\left(1,\ell(x;c)\right),\\
\ell(x;c)&=\frac{\log(1+x)}{\log(1+c)},\\
S(r;\tau)&=1-\exp\!\left(-\frac{r_+}{\tau}\right),\\
r_+&=\max(r,0).
\end{aligned}
\end{equation}

for the log-saturation and exponential-saturation functions used by the external scoring modules.

The following subsections specify anonymization, intrinsic rubrics and calibration, and the external source, sampling, and aggregation rules.

\subsection{Representative-Work Anonymization}

To reduce identity and reputation cues, HexEval maps each representative PDF $p$ to an anonymized Markdown document

\begin{equation}
\widetilde{p}=\mathcal{A}(p),
\end{equation}

where $\mathcal{A}$ denotes the anonymization and content-preservation procedure.
The PDF is converted to structured Markdown while retaining scientific content, including equations, tables,
figures, and captions. Rule-based filters remove names, affiliations, emails, acknowledgments, funding, and 
revealing citation metadata. An LLM cleaning stage removes residual institutional, group, project, and 
acknowledgment cues.

Methodological details, settings, formulations, results, limitations, and conclusions are retained. The same anonymized documents are supplied to D1--D3, which use independent prompts, rubrics, extraction procedures, and 
validation protocols.

To audit residual identity leakage, we sampled 100 representative works and used DeepSeek-V4-Flash as an 
automatic detector for explicit identity-revealing cues, including author names, affiliations, contact 
information, funding, acknowledgments, publication metadata, and identifying URLs. The residual leakage rate 
decreased from 0.53 after Stage 1 (MinerU conversion) to 0.52 after Stage 2 (regex-based cleaning), and further 
to 0 in this sampled audit after Stage 3 (LLM anonymization).

Anonymization mitigates but does not eliminate leakage: method names, datasets, benchmarks, writing style, or 
distinctive contributions may remain identifying. It is therefore a bias-mitigation mechanism, not a guarantee 
of identity-free evaluation.

\subsection{Intrinsic Research Quality Assessment}
\label{sec:intrinsic}

The intrinsic pathway receives only $\widetilde{\mathcal{P}}_s=\{\widetilde{p}_1,\ldots,\widetilde{p}_{n_s}\}$. For dimension $d\in\{1,2,3\}$ and paper $p$, the LLM returns a direct score $q_{d,p}$, subdimension scores, rationale, and diagnostic evidence. Identity, institution, venue, citations, and author metadata are excluded. The scholar-level direct score is the mean over valid works,

\begin{equation}
D_d^{\mathrm{direct}}(s)=
\frac{1}{|\mathcal{P}_s^{\mathrm{valid}}|}
\sum_{p\in\mathcal{P}_s^{\mathrm{valid}}}q_{d,p}.
\end{equation}

The three dimensions are not collapsed into one intrinsic score. With human labels, a dimension-specific Ridge calibrator is fitted for held-out validation. Let $\mathbf{x}_{d,p}$ contain the LLM subdimension and direct scores, and $y_{d,p}$ the human score. After standardization,

\begin{equation}
\widehat{\boldsymbol{\beta}}_d
=\arg\min_{\boldsymbol{\beta}}
\left\{
\left\|\mathbf{y}_d-\mathbf{X}_d\boldsymbol{\beta}\right\|_2^2
+\lambda_d\left\|\boldsymbol{\beta}\right\|_2^2
\right\},
\end{equation}

where $\lambda_d$ is selected on the calibration split. The test prediction is

\begin{equation}
\begin{aligned}
\widehat{y}_{d,p}&=\mathrm{clip}_{[1,4]}\\
&\quad\left(\widehat{\beta}_{d,0}+\mathbf{z}_{d,p}^{\top}
\widehat{\boldsymbol{\beta}}_d\right).
\end{aligned}
\end{equation}

The calibrator is used only when this fitted mapping is available; otherwise, the direct mean is retained. Thus, benchmark calibration is not presented as an unsupervised scoring rule.

\paragraph{D1: Research rigor.}
D1 evaluates whether claims are supported by sound methods and evidence. Its seven criteria are methodological validity, evidence adequacy, evaluation design, comparisons and controls, statistical or logical rigor, reproducibility and transparency, and limitation/claim calibration. The output contains criterion rationales and serious or minor weaknesses; unsupported claims reduce the relevant criterion rather than incur a reputation-based penalty.

\paragraph{D2: Methodological innovation.}
D2 evaluates originality in technical context. The main benchmark path receives extracted abstract, introduction, related-work/background, method, and conclusion sections from the anonymized paper, without identity or reputation cues. Its criteria are core originality, technical distinctiveness, nontriviality, and novelty-claim specificity. The evaluator distinguishes new mechanisms from new applications, tuning, implementation changes, and performance gains. An optional OpenAlex prior-work mode restricts candidates to works before the target year but is not used for the reported benchmark. The output includes the central method claim, novelty type, comparison rationale, and $q_{2,p}$.

\paragraph{D3: Scientific contribution.}
D3 measures scientific significance and usefulness rather than method novelty alone. Its criteria are problem importance, contribution substance, result value, and generality/reusability. The output contains the main contribution, strengths, serious and minor weaknesses, rationale, and $q_{3,p}$, allowing narrow but correct work to differ from broadly reusable work.

For all three dimensions, each saved score links to extracted paper content, subdimension values, and rationale; the scholar-level mean is therefore an aggregation of inspectable paper-level judgments.

\subsection{External Scholarly Behavior Assessment}
\label{sec:external}

The external pathway operates at the scholar level because these dimensions describe observable scholarly behavior rather than the quality of an individual paper. The evidence sources, attribution checks, and scoring rules are kept separate for each dimension.

\paragraph{D4: Knowledge translation.}
D4 measures validated translation of research into reusable software and patented
or otherwise documented intellectual property. Software evidence is collected
from GitHub and public project records. A community project contributes only when
its evidence confidence is strong or moderate, its repository is reachable, and
the scholar's contributor attribution is verified. Personal repositories are
restricted to owned, non-forked repositories with at least 100 stars. Patent
candidates are deduplicated into families; only strong or moderate, non-review-
required families receive positive weight. Weak, unverified, and review-required
items remain in the audit trail but do not contribute to the score.

For a repository $r$, the software impact is
\begin{equation}
\begin{aligned}
I_r&=0.60L(\mathrm{stars}_r;10000)\\
&\quad+0.30L(\mathrm{forks}_r;3000)+0.10a_r,
\end{aligned}
\end{equation}
where $a_r=1.0$ for activity within two years, $0.7$ for activity three to five
years old, $0.4$ for older or unknown activity, and $0.2$ for archived
repositories. A selected project receives
\begin{equation}
g_r=e_r w_r(1+3I_r),
\end{equation}
where the implemented evidence weights are $e_r=1.0$ for strong community
evidence, $0.6$ for moderate community evidence, and $0.45$ for a qualifying
personal repository. The type weights are $w_r=1.20$ for community projects,
$1.00$ for ordinary personal repositories, and $0.70$ for personal repositories
matching the low-value repository patterns. The top eight community projects and
top ten personal repositories are retained, and
$T_{\mathrm{soft}}(s)=S(G_s;12)$.

For a validated patent family $f$, the impact term is
\begin{equation}
I_f=0.50L(\mathrm{citations}_f;100)
 +0.25L(\mathrm{familySize}_f;10)+0.25o_f,
\end{equation}
where $o_f$ is the ownership score. The family score is
\begin{equation}
g_f=e_f(1.5b_f+3I_f),
\end{equation}
where $b_f=1.0$ for a granted family and $b_f=0.6$ otherwise, while
$e_f=1.0$ for strong evidence, $0.5$ for moderate evidence, and $0$ for weak
or review-required evidence. The positive family-score sum is saturated as
$T_{\mathrm{pat}}(s)=S(P_s;8)$. The final score is
\begin{equation}
D_4(s)=100[0.60T_{\mathrm{soft}}(s)+0.40T_{\mathrm{pat}}(s)].
\end{equation}
This score represents validated translation evidence rather than complete
individual contribution. Evidence strength, attribution status, validation
status, and source identifiers are retained for auditing.

\paragraph{D5: Research coherence.}
D5 estimates research coherence from a sparse chronological sample of a scholar's fuller career trajectory. We reuse 110 preselected computer-science scholars and retrieve their author-matched OpenAlex publication records. The full-career package is divided into five chronological bins. Papers enter the primary coherence corpus only when year, title, and abstract are all available; excluded records remain documented in the package audit trail.

The adjudicated reference is constructed from the full-career packages. GLM and DeepSeek independently score the five coherence dimensions: thematic consistency, temporal continuity, main-thread clarity, related-branch
integration, and low fragmentation. ChatGPT does not produce an independent score; it acts only as an anonymous judge of the two scorer outputs. The final reference value is the mean of the five final adjudicated dimension scores and is referred to as an adjudicated multi-LLM reference annotation, not as ground truth. A fixed 20/90 development/test split is used, and the test reference and sampling manifest are frozen before evaluation.

For the sparse evaluation, the same manifest supplies three papers per bin, or at most 15 papers per scholar, and each scholar is evaluated over five repeated samples. The evaluator receives only year, title, and abstract. If $c_{s,j}$ is the overall coherence score for repeat $j$, the reported prediction is

\begin{equation}
D_5(s)=\frac{1}{5}\sum_{j=1}^{5}c_{s,j},
\qquad
\sigma_5(s)=\operatorname{SD}(c_{s,1},\ldots,c_{s,5}).
\end{equation}

\paragraph{D6: Academic impact.}
D6 is an OpenAlex-based bibliometric anchor rather than a newly proposed composite index or a separately labeled benchmark. For each resolved scholar, we retrieve the OpenAlex author record and the author-matched work records. The canonical D6 value is the author-level h-index \cite{hirsch2005index} in {\tt authors.summary\_stats.h\_index}. If that field is unavailable, the implementation uses a documented h-index fallback computed from the retrieved
valid works.

Other OpenAlex fields are retained as auditable evidence but are not combined into an additional score. These fields include total citations, i10-index, works count, two-year mean citedness, FWCI, citation-normalized percentiles, recent citations and works, top works, pagination status, and the retrieval timestamp. Thus, D6 provides a reproducible and updateable citation-based impact signal, while D1--D5 capture non-bibliometric properties that h-index cannot represent.

\section{Experimental}
\subsection{Evaluation Dataset Construction}

\begin{table}[t]
\centering
\begin{tabular}{@{}ccll@{}}
\toprule
\textbf{Dim.} & \textbf{Size} & \textbf{Source} & \textbf{Reference} \\
\midrule
D1 & 300 & OpenReview & Soundness \\
D2 & 300 & OpenReview & Novelty \\
D3 & 300 & OpenReview & Contribution \\
D4 & 90  & Public records & Translation tier \\
D5 & 110 & OpenAlex & Coherence score \\
\bottomrule
\end{tabular}
\caption{Summary of the evaluation datasets for D1--D5.}
\label{tab:evaluation_datasets}
\end{table}

Because the dimensions use different evidence sources and validation roles, D1--D3 rely on human judgments, D4
uses curated scholarly evidence, D5 uses adjudicated multi-LLM references, and D6 uses the OpenAlex h-index
without a separately constructed benchmark.

\subsubsection{Intrinsic Quality Dataset.}
We use public OpenReview reviews with dimension-level human judgments. D1 uses NeurIPS soundness annotations,
D2 uses ICLR 2022 technical novelty and significance scores sampled by novelty quartiles, and D3 uses
contribution-related annotations.

\subsubsection{External Scholarly Behavior Dataset.}
D4--D5 are evaluated at scholar level because they characterize observable scholarly behavior rather than
paper quality. D4 groups scholars into high, middle, and low translation levels using awards and publicly
documented software/IP outcomes. D5 uses author-matched OpenAlex publication corpora for preselected computer-
science scholars. Full-career evidence packages are scored independently by GLM-5.2 \cite{zai2026glm52}, and DeepSeek-V4-Flash \cite{deepseek2026v4}, and GPT-5.5 \cite{openai2026gpt55} acts as an anonymous adjudicator of their outputs to produce the final coherence references. Evaluation uses sparse chronological publication samples.

\begin{table*}[ht]
\centering
\setlength{\tabcolsep}{1mm}
\begin{tabular}{@{}lccccccccc@{}}
\toprule
&
\multicolumn{3}{c}{\textbf{D1: Rigor}} &
\multicolumn{3}{c}{\textbf{D2: Innovation}} &
\multicolumn{3}{c}{\textbf{D3: Contribution}} \\
\cmidrule(lr){2-4}
\cmidrule(lr){5-7}
\cmidrule(lr){8-10}
\textbf{Method}
& $\rho \uparrow$ & \textbf{MAE} $\downarrow$ & \textbf{Acc.@0.5} $\uparrow$
& $\rho \uparrow$ & \textbf{MAE} $\downarrow$ & \textbf{Acc.@0.5} $\uparrow$
& $\rho \uparrow$ & \textbf{MAE} $\downarrow$ & \textbf{Acc.@0.5} $\uparrow$ \\
\midrule
Direct Overall
& .411 & .399 & \textbf{.73}
& .378 & .435 & .60
& \textbf{.291} & .516 & .54 \\

CoT
& .440 & .481 & .62
& .417 & .463 & \textbf{.70}
& .239 & .705 & .28 \\

Self-Reflection
& .402 & .510 & .64
& \textbf{.436} & .402 & .67
& .258 & .572 & .45 \\

Subdimension Mean
& .429 & .379 & .72
& .381 & .442 & .59
& .272 & .558 & .49 \\

\textbf{HexEval + Ridge}
& \textbf{.467} & \textbf{.361} & .70
& .356 & \textbf{.378} & .68
& .235 & \textbf{.295} & \textbf{.86} \\
\bottomrule
\end{tabular}
\caption{Agreement between intrinsic quality assessments and human peer-review
scores using Qwen3.6-27B. Higher values are better for Spearman's $\rho$ and
Acc.@0.5, whereas lower MAE is better. Bold values indicate the best result
within each dimension and metric.}
\label{tab:intrinsic_results}
\end{table*}

\subsection{Experimental Setting}
\label{sec:experimental_setting}

Local models are served with vLLM \cite{kwon2023efficient}. Representative PDFs are converted to structured Markdown using MinerU \cite{wang2024mineru}, processed by rule-based filters, and anonymized using Qwen2.5-72B-Instruct-AWQ \cite{yang2024qwen25} with temperature 0 and a maximum output length of 1,200 tokens.

All primary LLM-based predictions for D1--D3 and D5 use Qwen3.6-27B \cite{qwen2026qwen36} with a 32K context window. D1 uses temperature 0, D2 and D3 use temperature 0.1, and D5 uses temperature 0. The maximum output lengths are 3,000 tokens for D1, 1,800 tokens for D2 and D3, and 2,048 tokens for D5. Ridge calibration is trained only on the calibration split and evaluated on held-out test data. Model versions, prompts, retrieval dates, random seeds, and sampling manifests are recorded for reproducibility.

\paragraph{Baselines.}
For D1--D3, we compare HexEval with direct overall scoring, chain-of-thought prompting \cite{wei2022chain}, 
self-reflection prompting \cite{madaan2023selfrefine}, and the unweighted mean of structured subdimension 
scores. All methods use the same evaluator model, anonymized paper inputs, and dimension-specific scoring 
scales. HexEval additionally applies Ridge calibration trained only on the calibration split. 

For D4, we compare the full score with single-indicator and count-based baselines derived from the same verified software and patent evidence.

For D5, we compare HexEval-D5 with TF--IDF \cite{salton1988term}, SPECTER2 \cite{singh2023scirepeval}, and direct zero-shot LLM scoring. All D5 methods use the same frozen five-bin sampling manifest, three papers per bin, and five repeated samples.

\section{Results}
\subsection{Intrinsic Research Quality Evaluation}
\label{sec:intrinsic_evaluation}

We evaluate the three intrinsic dimensions by comparing model scores with human peer-review judgments. D1, D2, and D3 use reviewer-averaged rigor, technical novelty, and contribution scores as reference labels, 
respectively. Each dimension contains 100 held-out test papers disjoint from the calibration set. We fix the evaluator model to Qwen3.6-27B and compare direct overall scoring, chain-of-thought reasoning 
\cite{wei2022chain}, self-reflection \cite{madaan2023selfrefine}, simple averaging of structured subdimension scores, and HexEval with Ridge calibration. We report Spearman's rank correlation ($\rho$), mean absolute error (MAE), and the proportion of predictions within 0.5 points of the
human score (Acc@0.5).

\paragraph{D1: Research Rigor.}
HexEval performs strongest on rigor. Ridge calibration achieves the highest rank correlation ($\rho=.467$) and the lowest MAE (.361), improving over direct overall scoring on both measures. Direct scoring obtains the highest Acc@0.5 (.73), but its larger negative bias indicates less calibrated absolute scoring. Overall, structured rigor decomposition plus learned aggregation improves the consistency and scale alignment of soundness assessment.

\paragraph{D2: Methodological Innovation.}
Innovation remains the most difficult intrinsic dimension. Self-reflection gives the highest rank correlation ($\rho=.436$), and CoT gives the highest Acc@0.5 (.70). HexEval achieves the lowest MAE (.378), indicating better score-scale calibration, but it does not improve ordinal ranking. This suggests that calibration reduces systematic score error, while relative novelty ordering is still sensitive to model judgment.

\paragraph{D3: Scientific Contribution.}
For contribution, HexEval substantially improves absolute agreement: MAE drops from .516 under direct scoring to .295, and Acc@0.5 increases from .54 to .86. However, direct overall scoring still gives the highest rank correlation ($\rho=.291$). Thus, structured calibration is effective for aligning contribution scores with the human scale, but the relative ordering of papers remains challenging.

Overall, HexEval most clearly improves score calibration and absolute agreement across intrinsic dimensions. Ranking gains are strongest for D1, whereas D2 and D3 show that scale calibration and ordinal ranking are not always improved by the same mechanism.

\begin{figure*}[!ht]
\centering
\includegraphics[width=0.98\textwidth]{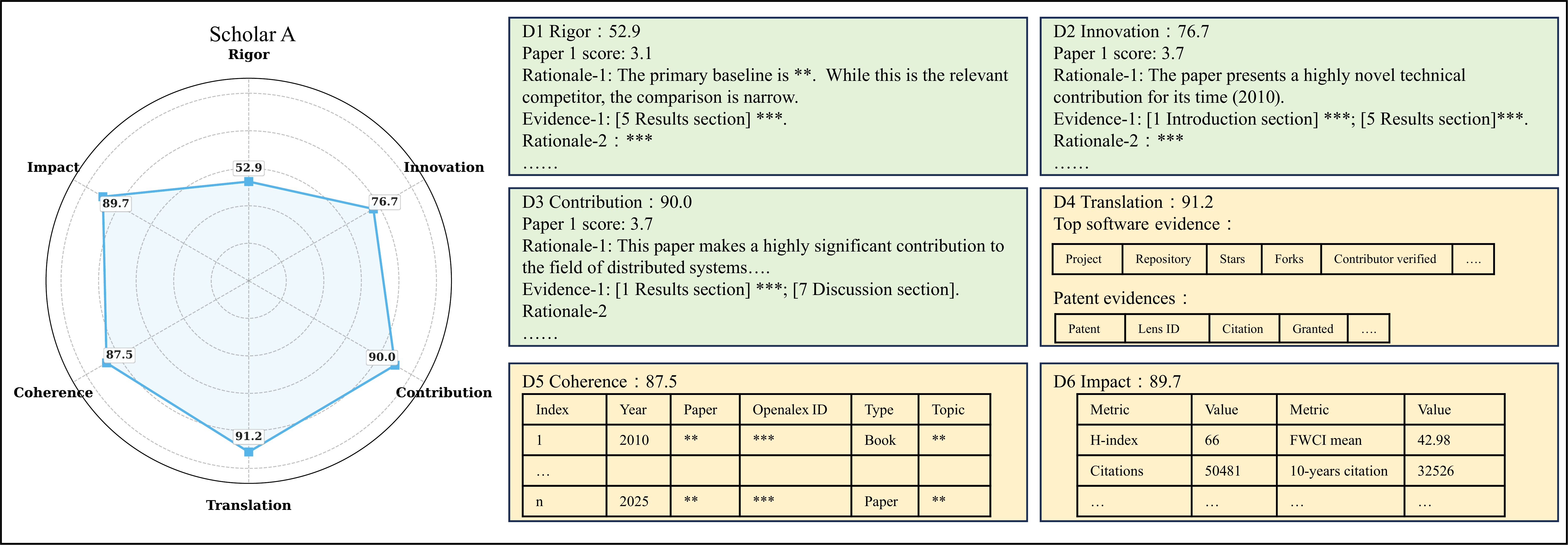}
\caption{End-to-end HexEval case study for Scholar A. The figure summarizes the six-dimensional profile and associated evidence outputs. D1--D3 use anonymized representative works, whereas D4--D6 use identity-resolved 
software, patent, and OpenAlex evidence. Identifying information is removed from the reported case study. The 
profile is illustrative and should be interpreted together with evidence coverage and attribution status.}
\label{fig:case_study}
\end{figure*}

\subsection{External Scholarly Behavior Evaluation}
\label{sec:external_evaluation}

We evaluate D4 against external knowledge-translation tiers and D5 against the frozen adjudicated coherence reference. D6 is implemented as a source-backed bibliometric indicator rather than evaluated against a 
separately constructed label set.

\paragraph{D4: Knowledge Translation.}
At the main thresholds of 30 and 50, the fixed D4 score achieves .611 accuracy
and .601 Macro-F1 on the balanced three-level benchmark. As shown in
Table~\ref{tab:d4_indicator_comparison}, the full D4 score obtains the highest
High--Low AUC (.930), exceeding GitHub max stars (.894) and raw
software-plus-patent counts (.893), the two strongest baselines. Its F1@$k$ and
Acc.@$k$ values are both .867, tying several simpler indicators. The principal
advantage of the complete D4 formulation therefore lies in separating high- and
low-translation scholars across the full ranking, rather than in improving
top-$k$ retrieval alone. The continuous ranking is independent of the reporting
thresholds, whereas the derived three-level classification remains sensitive to
the selected cut points; detailed threshold-sensitivity results are provided in
the appendix. Overall, these results support the joint use of verified software
and patent/IP evidence, while retaining evidence attribution, validation status,
and source identifiers for auditing.

\begin{table}[t]
\centering
{\small
\setlength{\tabcolsep}{1mm}
\begin{tabular}{@{}p{0.44\columnwidth}ccc@{}}
\toprule
\textbf{Method}
& \shortstack{\textbf{High--Low}\\\textbf{AUC}}
& \shortstack{\textbf{F1}\\\textbf{@$k$}}
& \shortstack{\textbf{Acc.}\\\textbf{@$k$}} \\
\midrule
Community project count
& .834 & .800 & .800 \\
Personal repository count
& .706 & \textbf{.867} & \textbf{.867} \\
Verified GitHub project count
& .848 & .833 & .833 \\
GitHub max stars
& .894 & \textbf{.867} & \textbf{.867} \\
Patent family count
& .617 & \textbf{.867} & \textbf{.867} \\
Patent citation count
& .693 & .833 & .833 \\
Raw software + patent counts
& .893 & .833 & .833 \\
\textbf{Full D4 score}
& \textbf{.930} & \textbf{.867} & \textbf{.867} \\
\bottomrule
\end{tabular}
}
\caption{Comparison of individual indicators and the full D4 score.}
\label{tab:d4_indicator_comparison}
\end{table}

\paragraph{D5: Research Coherence.}
As shown in Table~\ref{tab:d5_reference_comparison}, on the frozen
90-scholar test set, HexEval-D5 achieves $\rho=.743$, $\tau=.620$,
MAE $=.266$, and Acc.@0.5 $=.878$. Although direct zero-shot scoring
yields marginally higher rank correlations ($\rho=.746$, $\tau=.622$),
HexEval-D5 substantially improves absolute agreement, reducing MAE from
$.576$ to $.266$ and increasing Acc.@0.5 from $.489$ to $.878$.
TF--IDF and SPECTER2 obtain lower rank correlations, suggesting that
structured evaluation better aligns sparse publication samples with the
full-career coherence reference.

\begin{table}[t]
\centering
{\small
\setlength{\tabcolsep}{1mm}
\begin{tabular}{@{}lcccc@{}}
\toprule
\textbf{Method}
& \textbf{Spearman} $\uparrow$
& \textbf{Kendall} $\uparrow$
& \textbf{MAE} $\downarrow$
& \textbf{Acc.@0.5} $\uparrow$ \\
\midrule
TF--IDF
& .283 & .203 & .389 & .800 \\
SPECTER2
& .477 & .365 & .328 & .811 \\
Zero-shot LLM
& \textbf{.746} & \textbf{.622} & .576 & .489 \\
\textbf{HexEval-D5}
& .743 & .620 & \textbf{.266} & \textbf{.878} \\
\bottomrule
\end{tabular}
}
\caption{D5 sparse-to-full-career coherence evaluation on the frozen 90-scholar test set. All methods use the same five-bin, three-papers-per-bin sampling manifest and five repeats.}
\label{tab:d5_reference_comparison}
\end{table}

\paragraph{D6: Academic Impact.}
D6 uses the OpenAlex author-level h-index as its canonical impact indicator.
Total citations, i10-index, works count, two-year mean citedness, yearly citation
counts, FWCI, citation-normalized percentiles, recent activity, top works, and
retrieval and author-resolution metadata are retained as supporting evidence and
are not combined into a new composite score. Because D6 is a source-backed
operational indicator rather than a separately labeled benchmark, no D6
baseline-comparison or ablation table is reported.

\subsection{Case Study}

We illustrate the complete HexEval pipeline using an anonymized scholar, denoted as Scholar A. Six recent representative papers are anonymized and evaluated independently by D1--D3, while D4--D6 use identity-resolved
GitHub, Lens, and OpenAlex evidence. Identifying information is removed from the reported case study, while the underlying identity links are retained only for evidence attribution. Figure~\ref{fig:case_study} summarizes the
resulting six-dimensional profile and associated evidence outputs. The case study illustrates the reporting format and evidence flow rather than population-level accuracy.

D5 estimates research coherence from five repeated chronological samples of the author-resolved OpenAlex publication corpus. D6 uses the OpenAlex author-level h-index as its primary impact indicator, while other
bibliometric fields are retained only as supporting evidence. Both dimensions are interpreted together with evidence coverage and attribution status.

\subsection{Discussion}

HexEval is intended as an evidence-grounded decision-support framework rather than a replacement for expert judgment. Its main advantage is not a single superior ranking, but the separation of heterogeneous signals that conventional metrics collapse: D1--D3 characterize the quality of representative research, whereas D4--D6 describe knowledge translation, research coherence, and bibliometric impact. This separation improves interpretability, but the six dimensions should not be treated as interchangeable or mechanically averaged.
The framework remains limited by the completeness of public records, author disambiguation, and source-specific biases. GitHub and patent evidence may underrepresent some disciplines, OpenAlex coverage varies across fields, and the h-index remains sensitive to career age and citation practices. HexEval should therefore support, rather than determine, high-stakes assessment decisions.

\bibliography{aaai2027}

@article{hirsch2005index,
  author    = {Hirsch, Jorge E.},
  title     = {An index to quantify an individual's scientific research output},
  journal   = {Proceedings of the National Academy of Sciences},
  volume    = {102},
  number    = {46},
  pages     = {16569--16572},
  year      = {2005},
  doi       = {10.1073/pnas.0507655102}
}

@article{hicks2015leiden,
  author    = {Hicks, Diana and Wouters, Paul and Waltman, Ludo and de Rijcke, Sarah and R{\`a}fols, Ismael},
  title     = {Bibliometrics: The Leiden Manifesto for research metrics},
  journal   = {Nature},
  volume    = {520},
  number    = {7548},
  pages     = {429--431},
  year      = {2015},
  doi       = {10.1038/520429a}
}

@article{merton1968matthew,
  author    = {Merton, Robert K.},
  title     = {The Matthew Effect in Science: The reward and communication systems of science are considered},
  journal   = {Science},
  volume    = {159},
  number    = {3810},
  pages     = {56--63},
  year      = {1968},
  doi       = {10.1126/science.159.3810.56}
}

@article{thelwall2024chatgpt,
  author    = {Thelwall, Mike},
  title     = {Can {ChatGPT} evaluate research quality?},
  journal   = {Journal of Data and Information Science},
  volume    = {9},
  number    = {2},
  pages     = {1--21},
  year      = {2024},
  doi       = {10.2478/jdis-2024-0013}
}

@article{thelwall2025evaluating,
  author    = {Thelwall, Mike},
  title     = {Evaluating research quality with {Large Language Models}: An analysis of {ChatGPT's} effectiveness with different settings and inputs},
  journal   = {Journal of Data and Information Science},
  volume    = {10},
  number    = {1},
  pages     = {1--19},
  year      = {2025},
  doi       = {10.2478/jdis-2025-0011}
}

@article{priem2022openalex,
  author    = {Priem, Jason and Piwowar, Heather and Orr, Richard},
  title     = {{OpenAlex}: A fully-open index of scholarly works, authors, venues, institutions, and concepts},
  journal   = {arXiv preprint arXiv:2205.01833},
  year      = {2022},
  doi       = {10.48550/arXiv.2205.01833}
}

@article{radicchi2008universality,
  title={Universality of citation distributions: Toward an objective measure of scientific impact},
  author={Radicchi, Filippo and Fortunato, Santo and Castellano, Claudio},
  journal={Proceedings of the National Academy of Sciences},
  volume={105},
  number={45},
  pages={17268--17272},
  year={2008},
  publisher={National Acad Sciences}
}

@article{xie2021network,
  title={A network-embedding-based scholar assessment indicator},
  author={Xie, Zhe and Ouyang, ZhengZhen and Li, JianPing and Liu, GuoXiang and Hu, Xiao},
  journal={Journal of Informetrics},
  volume={15},
  number={3},
  pages={101171},
  year={2021},
  publisher={Elsevier}
}

@article{gorraiz2016individual,
  title={Individual bibliometric assessment at the University of Vienna},
  author={Gorraiz, Juan and Gumpenberger, Christian and Schloegl, Christian},
  journal={Bibliometrie - Praxis und Forschung},
  volume={5},
  year={2016}
}

@article{wainer2013correlations,
  title={What happens to computer science research after it is published?},
  author={Wainer, Jacques and Brahim, Gilberto Fridolin and Richard, Priscila},
  journal={Journal of the Association for Information Science and Technology},
  volume={64},
  number={5},
  pages={884--897},
  year={2013},
  publisher={Wiley Online Library}
}

@article{thelwall2023fields,
  title={Are citation indicators good alternatives to peer review? The case of the UK Research Excellence Framework},
  author={Thelwall, Mike},
  journal={Journal of Informetrics},
  volume={17},
  number={1},
  pages={101376},
  year={2023},
  publisher={Elsevier}
}

@inproceedings{pontika2022indicators,
  title={Indicators for Open Science and Responsible Research Assessment},
  author={Pontika, Natalia and Chatzopoulos, Stefania and Manola, Natalia and Manghi, Paolo},
  booktitle={Proceedings of the 26th International Conference on Science, Technology and Innovation Indicators (STI 2022)},
  year={2022},
  publisher={STI Conference}
}

@inproceedings{zhou2024reliable,
  title     = {Is {LLM} a Reliable Reviewer? A Comprehensive Evaluation of Large Language Models on Automatic Paper Reviewing Tasks},
  author    = {Zhou, Rui and Zeng, Xingshan and Wang, Junfeng and Yang, Yuhao and Liu, Juan},
  booktitle = {Proceedings of the 2024 Joint International Conference on Computational Linguistics, Language Resources and Evaluation (LREC-COLING)},
  year      = {2024}
}

@inproceedings{du2024critique,
  title     = {{LLMs} Assist {NLP} Researchers: Critique Paper (Meta-)Reviewing},
  author    = {Du, Jiangshu and Zhang, Yifan and Cao, Yixin and Fan, Zhihao and Yu, Zhiwei and Li, Yuhan and Wang, Benyou},
  booktitle = {Proceedings of the 2024 Conference on Empirical Methods in Natural Language Processing (EMNLP)},
  year      = {2024}
}

@inproceedings{jin2024agentreview,
  title     = {Exploring Peer Review Dynamics with {LLM} Agents},
  author    = {Jin, Yiqiao and Wu, Zhiyu and Tan, Zhen and Ge, Yuyang and Joty, Shafiq and Bing, Lidong},
  booktitle = {Proceedings of the 2024 Conference on Empirical Methods in Natural Language Processing (EMNLP)},
  year      = {2024}
}

@inproceedings{zhu2025deepreview,
  title     = {{DeepReview}: Improving {LLM}-Based Paper Review with Human-like Deep Thinking Process},
  author    = {Zhu, Minjun and Weng, Yixuan and Yang, Linyi and Zhang, Yue},
  booktitle = {Proceedings of the 63rd Annual Meeting of the Association for Computational Linguistics (ACL)},
  pages     = {29330--29355},
  year      = {2025},
  doi       = {10.18653/v1/2025.acl-long.1420}
}

@article{Thakkar2025CanLF,
  title={Can {LLM} feedback enhance review quality? A randomized study of 20K reviews at {ICLR} 2025},
  author={Nitya Thakkar and Mert Yuksekgonul and Jake Silberg and Animesh Garg and Nanyun Peng and Fei Sha and Rose Yu and Carl Vondrick and James Zou},
  journal={arXiv preprint arXiv:2504.09737},
  year={2025},
  url={https://arxiv.org/abs/2504.09737}
}

@misc{Kousha2025ChatGPT,
  title={Can {ChatGPT} evaluate research environments? Evidence from {REF2021}},
  author={Kayvan Kousha and Mike Thelwall and Elizabeth Gadd},
  year={2025},
  month={December},
  eprint={2512.05202},
  archivePrefix={arXiv},
  primaryClass={cs.DL},
  url={https://arxiv.org/abs/2512.05202}
}

@misc{Thelwall2024InWhich,
  title={In which fields can {ChatGPT} detect journal article quality? An evaluation of {REF2021} results},
  author={Mike Thelwall and Abdallah Yaghi},
  year={2024},
  month={September},
  eprint={2409.16695},
  archivePrefix={arXiv},
  primaryClass={cs.DL},
  url={https://arxiv.org/abs/2409.16695}
}

@article{nunkoo2026globalsouth,
  title={A Global South Strategy for Evaluating Research Value with ChatGPT},
  author={Nunkoo, Robin and Thelwall, Mike},
  journal={Quantitative Science Studies},
  year={2026},
  doi={10.1162/QSS.a.460},
  url={https://arxiv.org/abs/2508.01882}
}

@inproceedings{lewis2020retrieval,
  title     = {Retrieval-Augmented Generation for Knowledge-Intensive NLP Tasks},
  author    = {Lewis, Patrick and Perez, Ethan and Piktus, Aleksandra and
               Petroni, Fabio and Karpukhin, Vladimir and Goyal, Naman and
               K{\"u}ttler, Heinrich and Lewis, Mike and Yih, Wen-tau and
               Rockt{\"a}schel, Tim and Riedel, Sebastian and Kiela, Douwe},
  booktitle = {Advances in Neural Information Processing Systems},
  volume    = {33},
  pages     = {9459--9474},
  year      = {2020}
}

@inproceedings{asai2022evidentiality,
  title     = {Evidentiality-Guided Generation for Knowledge-Intensive NLP Tasks},
  author    = {Asai, Akari and Gardner, Matt and Hajishirzi, Hannaneh},
  booktitle = {Findings of the Association for Computational Linguistics:
               NAACL 2022},
  year      = {2022}
}

@inproceedings{wadden2020fact,
  title     = {Fact or Fiction: Verifying Scientific Claims},
  author    = {Wadden, David and Lin, Shanchuan and Lo, Kyle and Wang, Lucy Lu
               and van Zuylen, Madeleine and Cohan, Arman and Hajishirzi,
               Hannaneh},
  booktitle = {Proceedings of the 2020 Conference on Empirical Methods in
               Natural Language Processing},
  pages     = {7534--7550},
  year      = {2020}
}

@inproceedings{gao2023rarr,
  title     = {RARR: Researching and Revising What Language Models Say,
               Using Language Models},
  author    = {Gao, Luyu and Dai, Zhuyun and Pasupat, Panupong and Chen,
               Anthony and Chaganty, Arun Tejasvi and Fan, Yicheng and Zhao,
               Vincent and Lao, Ni and Lee, Hongrae and Juan, Da-Cheng and
               Guu, Kelvin},
  booktitle = {Proceedings of the 61st Annual Meeting of the Association for
               Computational Linguistics},
  year      = {2023}
}

@inproceedings{wadden-etal-2022-scifact,
    title = "{S}ci{F}act-Open: Towards open-domain scientific claim verification",
    author = "Wadden, David  and
      Lo, Kyle  and
      Kuehl, Bailey  and
      Cohan, Arman  and
      Beltagy, Iz  and
      Wang, Lucy Lu  and
      Hajishirzi, Hannaneh",
    editor = "Goldberg, Yoav  and
      Kozareva, Zornitsa  and
      Zhang, Yue",
    booktitle = "Findings of the Association for Computational Linguistics: EMNLP 2022",
    month = dec,
    year = "2022",
    address = "Abu Dhabi, United Arab Emirates",
    publisher = "Association for Computational Linguistics",
    url = "https://aclanthology.org/2022.findings-emnlp.347/",
    doi = "10.18653/v1/2022.findings-emnlp.347",
    pages = "4719--4734",
}

@inproceedings{jacovi2020faithfully,
  title     = {Towards Faithfully Interpretable NLP Systems:
               How Should We Define and Evaluate Faithfulness?},
  author    = {Jacovi, Alon and Goldberg, Yoav},
  booktitle = {Proceedings of the 58th Annual Meeting of the
               Association for Computational Linguistics},
  pages     = {4198--4205},
  year      = {2020},
  doi       = {10.18653/v1/2020.acl-main.386}
}

@inproceedings{kwon2023efficient,
  title     = {Efficient Memory Management for Large Language Model Serving
               with {PagedAttention}},
  author    = {Kwon, Woosuk and Li, Zhuohan and Zhuang, Siyuan and
               Sheng, Ying and Zheng, Lianmin and Yu, Cody Hao and
               Gonzalez, Joseph E. and Zhang, Hao and Stoica, Ion},
  booktitle = {Proceedings of the 29th Symposium on Operating Systems Principles},
  pages     = {611--626},
  year      = {2023},
  doi       = {10.1145/3600006.3613165}
}

@article{wang2024mineru,
  title   = {{MinerU}: An Open-Source Solution for Precise Document
             Content Extraction},
  author  = {Wang, Bin and Xu, Chao and Zhao, Xiaomeng and Ouyang, Linke
             and Wu, Fan and Zhao, Zhiyuan and Xu, Rui and Liu, Kaiwen
             and Qu, Yuan and Shang, Fukai and Zhang, Bo and Wei, Liqun
             and Sui, Zhihao and Li, Wei and Shi, Botian and Qiao, Yu
             and Lin, Dahua and He, Conghui},
  journal = {arXiv preprint arXiv:2409.18839},
  year    = {2024},
  doi     = {10.48550/arXiv.2409.18839}
}

@article{yang2024qwen25,
  title   = {{Qwen2.5} Technical Report},
  author  = {Yang, An and Yang, Baosong and Zhang, Beichen and Hui, Binyuan
             and Zheng, Bo and Yu, Bowen and Li, Chengyuan and Liu, Dayiheng
             and Huang, Fei and Wei, Haoran and Lin, Huan and Yang, Jian
             and Tu, Jianhong and Zhang, Jianwei and Yang, Jianxin
             and Yang, Jiaxi and Zhou, Jingren and Lin, Junyang and Dang, Kai
             and Lu, Keming and Bao, Keqin and Yang, Kexin and Yu, Le
             and Li, Mei and Xue, Mingfeng and Zhang, Pei and Zhu, Qin
             and Men, Rui and Lin, Runji and Li, Tianhao and Xia, Tingyu
             and Ren, Xingzhang and Ren, Xuancheng and Fan, Yang and Su, Yang
             and Zhang, Yichang and Wan, Yu and Liu, Yuqiong and Cui, Zeyu
             and Zhang, Zhenru and Qiu, Zihan},
  journal = {arXiv preprint arXiv:2412.15115},
  year    = {2024},
  doi     = {10.48550/arXiv.2412.15115}
}

@misc{qwen2026qwen36,
  title  = {{Qwen3.6-27B}: Flagship-Level Coding in a {27B} Dense Model},
  author = {{Qwen Team}},
  month  = {April},
  year   = {2026},
  url    = {https://qwen.ai/blog?id=qwen3.6-27b}
}

@inproceedings{wei2022chain,
  title     = {Chain-of-Thought Prompting Elicits Reasoning in Large
               Language Models},
  author    = {Wei, Jason and Wang, Xuezhi and Schuurmans, Dale
               and Bosma, Maarten and Ichter, Brian and Xia, Fei
               and Chi, Ed H. and Le, Quoc V. and Zhou, Denny},
  booktitle = {Advances in Neural Information Processing Systems},
  volume    = {35},
  pages     = {24824--24837},
  year      = {2022}
}

@inproceedings{madaan2023selfrefine,
  title     = {Self-Refine: Iterative Refinement with Self-Feedback},
  author    = {Madaan, Aman and Tandon, Niket and Gupta, Prakhar
               and Hallinan, Skyler and Gao, Luyu and Wiegreffe, Sarah
               and Alon, Uri and Dziri, Nouha and Prabhumoye, Shrimai
               and Yang, Yiming and Gupta, Shashank
               and Majumder, Bodhisattwa Prasad and Hermann, Katherine
               and Welleck, Sean and Yazdanbakhsh, Amir and Clark, Peter},
  booktitle = {Advances in Neural Information Processing Systems},
  volume    = {36},
  pages     = {46534--46594},
  year      = {2023}
}

@article{salton1988term,
  title   = {Term-Weighting Approaches in Automatic Text Retrieval},
  author  = {Salton, Gerard and Buckley, Christopher},
  journal = {Information Processing \& Management},
  volume  = {24},
  number  = {5},
  pages   = {513--523},
  year    = {1988},
  doi     = {10.1016/0306-4573(88)90021-0}
}

@inproceedings{singh2023scirepeval,
  title     = {{SciRepEval}: A Multi-Format Benchmark for Scientific
               Document Representations},
  author    = {Singh, Amanpreet and D'Arcy, Mike and Cohan, Arman
               and Downey, Doug and Feldman, Sergey},
  booktitle = {Proceedings of the 2023 Conference on Empirical Methods
               in Natural Language Processing},
  pages     = {5548--5566},
  address   = {Singapore},
  publisher = {Association for Computational Linguistics},
  year      = {2023},
  doi       = {10.18653/v1/2023.emnlp-main.338}
}

@misc{zai2026glm52,
  author       = {{Z.ai}},
  title        = {{GLM-5.2}: Built for Long-Horizon Tasks},
  year         = {2026},
  month        = jun,
  howpublished = {\url{https://z.ai/blog/glm-5.2}},
  note         = {Accessed: 2026-07-24}
}

@misc{deepseek2026v4,
  author       = {{DeepSeek-AI}},
  title        = {{DeepSeek V4 Preview Release}},
  year         = {2026},
  month        = apr,
  howpublished = {\url{https://api-docs.deepseek.com/news/news260424/}},
  note         = {Accessed: 2026-07-24}
}

@misc{openai2026gpt55,
  author       = {{OpenAI}},
  title        = {Introducing {GPT-5.5}},
  year         = {2026},
  month        = apr,
  howpublished = {\url{https://openai.com/index/introducing-gpt-5-5/}},
  note         = {Accessed: 2026-07-24}
}


\end{document}